\documentclass[mlmain]{jmlr}

\usepackage{xcolor}
\usepackage{tikz}
\usetikzlibrary{positioning}
\usepackage{booktabs}
\usepackage[load-configurations=version-1]{siunitx}

\jmlrvolume{}
\firstpageno{1}
\editors{}
\jmlryear{2026}
\jmlrworkshop{Symmetry and Geometry in Neural Representations}
\jmlrproceedings{PMLR}{Proceedings of Machine Learning Research}

\makeatletter
\AtBeginDocument{\def\@jmlrproceedings{}}
\makeatother

\title[Geometric Invariant of Attention]{A Geometric Invariant of Attention Predicts Head Specialization from Weights Alone}

\author{%
  \Name[Bento]{Miguel P. Bento\thanks{Accepted as an extended abstract at the Symmetry and Geometry in Neural Representations Workshop (NeurReps), NeurIPS 2026. This work was conducted independently of the author's role at Feedzai.}} \Email{miguel.pedra@feedzai.com}\\
  \addr Feedzai, Lisbon, Portugal
}

\begin{document}

\maketitle

\begin{abstract}
Transformers are built with continuous symmetries in the attention mechanism. We derive $\rho$, a scalar invariant that measures each head's capacity to couple to rotary positional encodings. This framework allows us to analyze heads without forward passes, opening a new research direction into data-independent analysis of modern LLMs. Empirically, we find that $\rho$ predicts induction heads, the most-studied circuit type in mechanistic interpretability, with Spearman correlation between $-0.55$ and $-0.74$ across four bias-free RoPE models spanning $135$M to $7$B parameters. As a preliminary proof of concept, we use $\rho$ to apply YaRN to only the top-50\% positional heads, which degrades perplexity less than uniform application. Code is available at \url{https://github.com/MPedraBento/rho-invariant}.
\end{abstract}


\section{Introduction}
\label{sec:intro}

Mechanistic interpretability is grounded in the meaning behind transformer head activations. Some approaches are activation patching \citep{conmy2023towards}, probing classifiers, or behavioral diagnostics on crafted inputs \citep{olsson2022context}. Recent work connecting RoPE to heads relies heavily on data \citep{Barbero2024RoundAR}. In this paper we ask: how much can we learn from the weights alone? Analogously to physics, where theories often have fewer observables than parameters, the attention mechanism admits basis-invariant quantities. These, in turn, characterize heads independently of the choice of basis.

The attention mechanism works by using quadratic forms in the query-key space $G_{QK} = W_Q^\top W_K \in \mathbb{R}^{d \times d}$ and the value-output space $G_{VO} = W_O W_V \in \mathbb{R}^{d \times d}$. The static composite $G_{QK}(0)=W_Q^\top W_K$ is invariant under the full head-space reparameterization $(W_Q,W_K)\mapsto(M^{-\top}W_Q, M W_K)$ for any $M\in GL(d_h)$. In RoPE-based models the rotation $R(\Delta)$ is block-diagonal with one $2\times 2$ rotation per frequency plane, and requiring invariance of the whole family $G_{QK}(\Delta)$ breaks $GL(d_h)$ down to the group $(\mathbb{C}^*)^{d_h/2}\cong(SO(2)\times\mathbb{R}_{>0})^{d_h/2}$, such that within each plane a rotation $R$ and a positive rescaling $D$ act as $W_Q\mapsto D^{-1}R\,W_Q$, $W_K\mapsto D R\,W_K$ leaving $G_{QK}(\Delta)$ unchanged. The form $G_{QK}$ is thus invariant under this residual symmetry, making it a basis-independent observable of the head's structure \citep{elhage2021mathematical}.

\section{Theory}
\label{sec:theory}

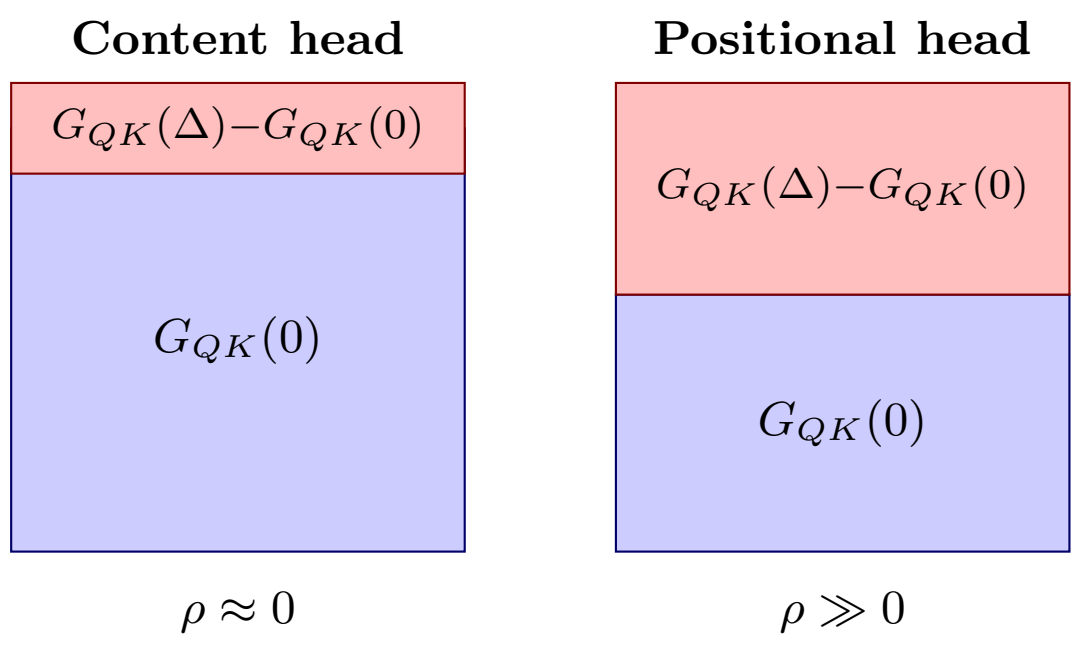
\begin{figure}[t]
\centering
\begin{tikzpicture}[font=\footnotesize]

\node[font=\footnotesize\bfseries] at (1.5, 3.4) {Content head};

\fill[blue!20, draw=blue!40!black] (0, 0) rectangle (3, 2.8);
\node at (1.5, 1.4) {$G_{QK}(0)$};

\fill[red!25, draw=red!50!black] (0, 2.5) rectangle (3, 3.1);
\node[font=\scriptsize] at (1.5, 2.8) {$G_{QK}(\Delta){-}G_{QK}(0)$};

\node at (1.5, -0.4) {$\rho \approx 0$};

\node[font=\footnotesize\bfseries] at (5.5, 3.4) {Positional head};

\fill[blue!20, draw=blue!40!black] (4, 0) rectangle (7, 1.7);
\node at (5.5, 0.85) {$G_{QK}(0)$};

\fill[red!25, draw=red!50!black] (4, 1.7) rectangle (7, 3.1);
\node[font=\scriptsize] at (5.5, 2.4) {$G_{QK}(\Delta){-}G_{QK}(0)$};

\node at (5.5, -0.4) {$\rho \gg 0$};

\end{tikzpicture}
\caption{The $\rho$ invariant measures the relative magnitude of the positional perturbation to the content term. Content heads (left) are dominated by $G_{QK}(0)$; positional heads (right) have a large position-dependent component.}
\label{fig:diagram}
\end{figure}

For models using Rotary Position Encoding \citep[RoPE;][]{SU2024127063}, the attention logit decomposes as
\begin{equation}\label{eq:rope_logit}
s_{ij} = x_i^\top \, G_{QK}(\Delta) \, x_j = x_i^\top \, G_{QK}(0) \, x_j + x_i^\top \, P(\Delta) \, x_j \, ,
\end{equation}
where $G_{QK}(0) = W_Q^\top W_K$ is the static term and $P(\Delta) = W_Q^\top [R(\Delta) - I] W_K$ is the positional perturbation induced by the RoPE rotation $R(\Delta)$ at relative position $\Delta = i - j$, with inputs $x_i \in \mathbb{R}^d$ and matrices $W_{Q,K} \in \mathbb{R}^{d_h \times d}$, so that $G_{QK} = W_Q^\top W_K \in \mathbb{R}^{d \times d}$.

We define a positional capacity invariant as a measure of the capacity a head has to couple to positional encodings. We can define a simple basis-invariant scalar that quantifies how much the bilinear form varies with position,
\begin{equation}\label{eq:rho}
\rho = \frac{1}{|\mathcal{D}|} \sum_{\Delta \in \mathcal{D}} \frac{\|P(\Delta)\|_F^2}{\|G_{QK}(0)\|_F^2}, \quad \mathcal{D} = \{2^n\}_{n=0}^{\lfloor\log_2 L\rfloor} \, ,
\end{equation}
where $L$ is the context length. The sampling set $\mathcal{D}$ is logarithmically spaced because the RoPE frequencies $\omega_k = \theta^{-2k/d_h}$ are geometrically spaced, so uniform sampling in $\log\Delta$ probes all frequency bands up to the context horizon $L$. Thus, $\rho$ measures positional capacity over the sampled range, not at arbitrarily long range.

The invariant is non-negative $\rho \geq 0$, with equality iff $G_{QK}(\Delta) = G_{QK}(0)$ for all $\Delta \in \mathcal{D}$. Being a function of $G_{QK}$, $\rho$ is additionally scale-invariant, unchanged under $W_Q \to \alpha W_Q$ or $W_K \to \alpha W_K$. It is, however, sensitive to the softmax shift $s_{ij} \to s_{ij} + c_i$, which leaves attention patterns unchanged; in practice weight decay penalizes $\|W_Q\|$ and $\|W_K\|$, discouraging large components in $G_{QK}$ that do not affect the output, thus mitigating this issue.

\begin{proposition}[$\rho=0$ gives position-independent logits on $\mathcal{D}$]
\label{prop:rho_zero}
Fix a head with static term $G_{QK}(0)=W_Q^\top W_K$ and assume $G_{QK}(0)\neq\mathbf{0}$. Let $\mathcal{D}$ be any (nonempty) set of relative positions. Then $\rho=0$ if and only if $P(\Delta)=\mathbf{0}$ for every $\Delta\in\mathcal{D}$, in which case the attention logit
\begin{equation}
s_{ij}=x_i^\top G_{QK}(\Delta)\,x_j=x_i^\top G_{QK}(0)\,x_j
\end{equation}
is independent of the relative position $\Delta=i-j$ for all $\Delta\in\mathcal{D}$ and all residual-stream inputs $x_i,x_j$.
\end{proposition}
\begin{proof}
By \eqref{eq:rho}, $\rho$ is a sum of terms $\|P(\Delta)\|_F^2/\|G_{QK}(0)\|_F^2$, each a squared Frobenius norm over the common denominator $\|G_{QK}(0)\|_F^2>0$ (positive by the assumption $G_{QK}(0)\neq\mathbf{0}$), hence each $\ge 0$. A sum of non-negative terms vanishes iff every term vanishes, so $\rho=0$ iff $P(\Delta)=\mathbf{0}$ for every $\Delta\in\mathcal{D}$. Substituting $P(\Delta)=G_{QK}(\Delta)-G_{QK}(0)=\mathbf{0}$ into \eqref{eq:rope_logit} gives $s_{ij}=x_i^\top G_{QK}(0)\,x_j$, which contains no dependence on $\Delta$.
\end{proof}

Crucially, $\rho$ measures \emph{capacity}, not usage, as a high-$\rho$ head may still ignore position on a given input, but a head with $\rho=0$ produces logits with no positional dependence across $\mathcal{D}$, for any input (Proposition~\ref{prop:rho_zero}).

\section{Results}

\subsection{$\rho$ predicts induction heads without data}

Induction heads \citep{olsson2022context} implement the ``A B \ldots A $\to$ B'' copying pattern, which is the most-studied circuit in mechanistic interpretability. We measure induction scores via repeated random sequences and correlate with $\rho$.

We find that induction heads have extremely low $\rho$, consistently across four
bias-free RoPE models spanning $135$M to $7$B parameters (Table~\ref{tab:induction}).
The Spearman correlation between $\rho$ and the induction score is negative and highly
significant in every model ($r$ from $-0.55$ to $-0.74$, all $p < 10^{-40}$), and the
ten strongest induction heads always lie in the low-$\rho$ tail (bottom $7$--$30\%$ of
the $\rho$ distribution).

\begin{table}[t]
\floatconts
  {tab:induction}
  {\caption{$\rho$ vs.\ induction score across four bias-free RoPE models. Spearman
  $r$ (all $p<10^{-40}$) and the smallest $\rho$-percentile band containing the top-10
  induction heads.}}
  {\begin{tabular}{lccc}
  \toprule
  \bfseries Model & \bfseries Params & \bfseries Spearman $r$ & \bfseries top-10 within \\
  \midrule
  SmolLM2-135M      & 135M & $-0.74$ & bottom 14\% \\
  TinyLlama-1.1B    & 1.1B & $-0.71$ & bottom 7\% \\
  SmolLM2-1.7B      & 1.7B & $-0.55$ & bottom 17\% \\
  Mistral-7B        & 7B   & $-0.59$ & bottom 30\% \\
  \bottomrule
  \end{tabular}}
\end{table}

This result supports our hypothesis of capacity interpretation. Induction heads perform content matching, a position-invariant operation. The low $\rho$ reflects minimal coupling to RoPE, thus ensuring activation-independent decoupling. The correlation is imperfect because induction is a multi-head circuit. Thus, low $\rho$ is necessary but not sufficient for induction, precisely the capacity interpretation.

\begin{figure}[t]
\floatconts
  {fig:scatter}
  {\caption{$\rho$ vs.\ induction score for all heads in TinyLlama-1.1B. Induction heads cluster in the low-$\rho$ regime.}}
  {\includegraphics[width=0.55\linewidth]{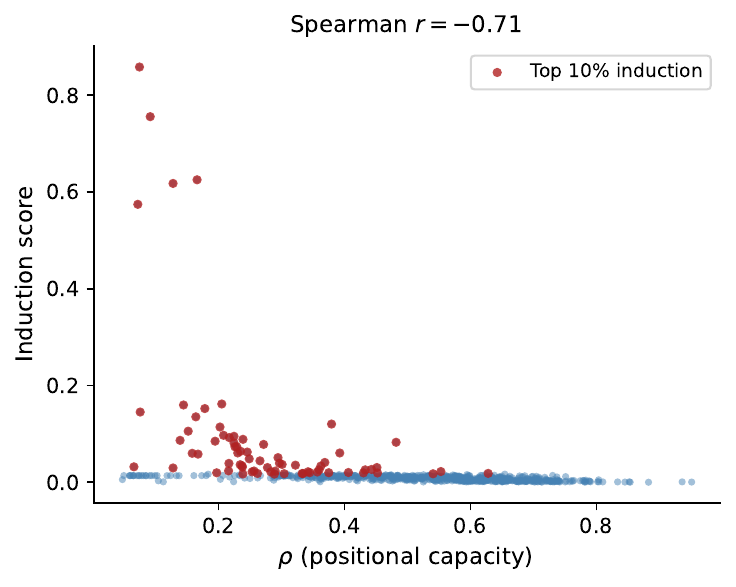}}
\end{figure}

\subsection{Application: capacity-gated context extension}

Uniform RoPE scaling methods \citep{Chen2023ExtendingCW} degrade short-context performance because they modify all heads. We apply YaRN \citep{peng2024yarn} to the top-50\% heads by $\rho$, leaving the remaining heads at the original positions.

\begin{table}[t]
\floatconts
  {tab:yarn}
  {\caption{Context extension at $S{=}4\times$ (Mistral-7B). Targeted applies YaRN only to the top-50\% $\rho$ heads.}}
  {\begin{tabular}{lcc}
  \toprule
  \bfseries Method & \bfseries Short PPL (512) & \bfseries Long PPL (4096) \\
  \midrule
  Baseline ($S{=}1$) & 7.63 & 5.44 \\
  YaRN (all heads) & 19.37 & 162.96 \\
  YaRN + Targeted (ours) & \textbf{12.92} & \textbf{130.34} \\
  \bottomrule
  \end{tabular}}
\end{table}

The targeted variant degrades perplexity less than uniform YaRN at both the short and the long context, as seen in Table~\ref{tab:yarn}. The evaluation stays within the model's native context. The inflated long context values are due to the evaluation near the context limit.

\section{Discussion and limitations}

We have shown that a single basis-invariant scalar, computable from weights alone, predicts induction heads. The key contribution is the \emph{capacity} interpretation. $\rho$ measures what a head's geometry allows it to couple to, not what it does on any given input.

This framework generalizes beyond the $\rho$ invariant. The full set of invariant quantities ($G_{QK}$ eigenspectrum, $G_{VO}$ singular values, the commutator $\|[G_{QK}, G_{VO}]\|_F$) defines an invariant signature per head. Future work entails a more comprehensive study of the weight invariants of transformers.

Finally, we validate on four bias-free RoPE models up to 7B parameters. The YaRN+Targeted application is a preliminary proof of concept on Mistral-7B, demonstrated within the native context window. Further evaluation is left to future work. The framework applies specifically to RoPE-based architectures, and to the QK sector.

\bibliography{references}

\end{document}